\documentclass[10pt,twocolumn,letterpaper]{article}

\usepackage{iccv}              

\definecolor{iccvblue}{rgb}{0.21,0.49,0.74}
\usepackage[pagebackref,breaklinks,colorlinks,allcolors=iccvblue]{hyperref}
\usepackage{url}            
\usepackage{booktabs}       
\usepackage{amsfonts}       
\usepackage{nicefrac}       
\usepackage{microtype}      
\usepackage{xcolor}         
\usepackage{graphicx}
\usepackage{multirow}
\usepackage{amssymb}
\usepackage{amsmath}
\usepackage{pifont}
\usepackage{color} 
\usepackage{soul} 
\newcommand{\cmark}{\ding{51}}%
\newcommand{\xmark}{\ding{55}}%

\def\paperID{10138} 
\def\confName{ICCV}
\def\confYear{2025}

\title{3DGraphLLM: Combining Semantic Graphs and Large Language Models for 3D Scene Understanding}

\author{
Tatiana Zemskova\textsuperscript{1,2} \quad
Dmitry Yudin\textsuperscript{1,2} \\
\textsuperscript{1}AIRI,
\textsuperscript{2}MIPT \\
}

\begin{document}
\maketitle

\begin{abstract}
A 3D scene graph represents a compact scene model by capturing both the objects present and the semantic relationships between them, making it a promising structure for robotic applications. To effectively interact with users, an embodied intelligent agent should be able to answer a wide range of natural language queries about the surrounding 3D environment. Large Language Models (LLMs) are beneficial solutions for user-robot interaction due to their natural language understanding and reasoning abilities. Recent methods for learning scene representations have shown that adapting these representations to the 3D world can significantly improve the quality of LLM responses. However, existing methods typically rely only on geometric information, such as object coordinates, and overlook the rich semantic relationships between objects. In this work, we propose 3DGraphLLM, a method for constructing a learnable representation of a 3D scene graph that explicitly incorporates semantic relationships. This representation is used as input to LLMs for performing 3D vision-language tasks. In our experiments on popular ScanRefer, Multi3DRefer, ScanQA, Sqa3D, and Scan2cap datasets, we demonstrate that our approach outperforms baselines that do not leverage semantic relationships between objects. The code is publicly available at \url{https://github.com/CognitiveAISystems/3DGraphLLM}.
\end{abstract}

\section{Introduction}


In this paper, we consider scene understanding in the context of 3D vision-language tasks: 3D referred object grounding task, 3D dense scene captioning and 3D visual question answering. The 3D referred object grounding task involves identifying a region within a 3D scene that corresponds to a natural language query. These queries often describe object properties (e.g., color, size) as well as spatial relationships (e.g., a mug on a table). A common setup of this problem assumes access to a 3D reconstruction of the scene, such as a point cloud, mesh, or NeRF. The objective is to predict the bounding box of the object or region referenced in the query. The goal of 3D dense scene captioning is to generate a textual description of a selected object in the 3D scene, including its attributes or relationships. Finally, the goal of the 3D visual question answering task is to generate text answers to various questions about the properties of the scene. It seems promising to explicitly use a three-dimensional scene graph to solve these tasks.

\begin{figure}[t]
  \centering
  \includegraphics[width=\linewidth]{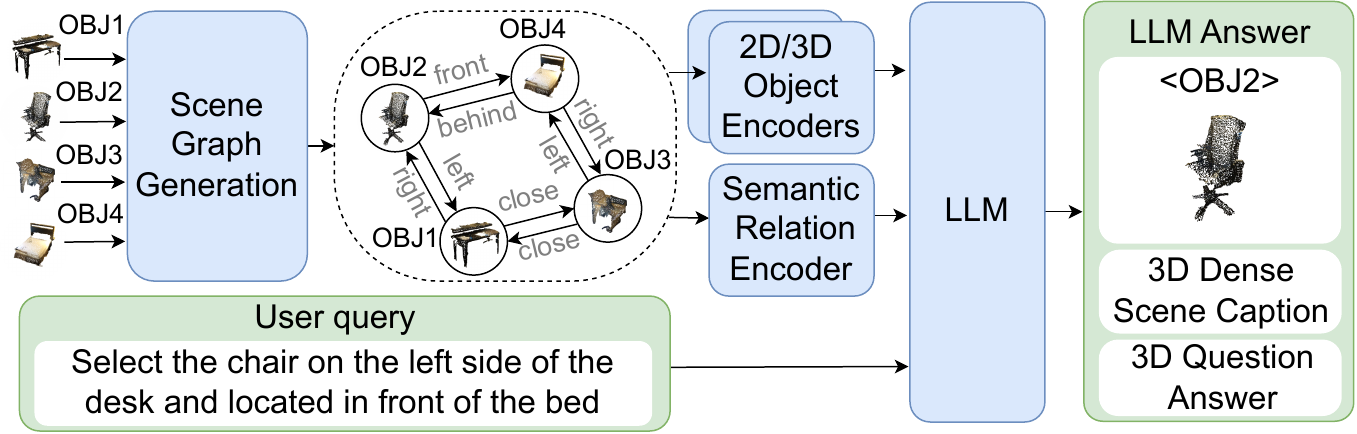}
  \caption{\label{ga}The proposed \textit{3DGraphLLM} approach leverages 3D semantic scene graph learnable representation supplied as input to an LLM to perform various 3D vision-language tasks.}
\end{figure}


A 3D scene graph provides a unified representation of a scene by storing multimodal information about individual objects, along with their semantic relationships~\citep{wang2023vl, koch2024open3dsg} and hierarchical organization~\citep{werby2024hierarchical, honerkamp2024language}. It also supports real-time updates in dynamic environments, making it suitable for interactive scenes~\citep{ rosinol2021kimera, ozsoy2023labrad}. Furthermore, representing the scene as a graph enables the use of graph algorithms for tasks such as navigation~\citep{zhou2023optimal, he2024relation, honerkamp2024language} and object search based on textual queries~\citep{feng2021free, chang2023context, werby2024hierarchical, gu2024conceptgraphs}.

Solving 3D vision-language tasks is essential for embodied intelligent agents~\citep{chen2020scanrefer, chen2021scan2cap, azuma2022scanqa}. To interact effectively with users, such agents must be able to describe their environment and answer questions about its properties using natural language. 
Large Language Models (LLMs) are particularly well-suited for this, thanks to their strong capabilities in language understanding and commonsense reasoning. They can interpret user queries and match them to objects in a scene, even when the queries are vague or indirect~\citep{hong20233d, wang2024large, gu2024conceptgraphs}.
By leveraging LLMs, it becomes easier to adapt the method to new object categories and relationships mentioned in referring expressions. LLMs can also handle complex queries that describe an object by its function rather than its name (e.g., "somewhere to sit").


A 3D scene can be represented for input to an LLM either as text~\citep{gu2024conceptgraphs, linok2024beyond, werby2024hierarchical, honerkamp2024language, yang2024llm, yuan2024visual} or as a implicit learnable representation~\citep{hong20233d, chen2023ll3da, huang2023chat, chen2024grounded, cheng2024spatialrgpt}. Learnable representations encode objects and their relationships into embeddings, using significantly fewer tokens than textual descriptions. This compact form not only increases the speed of LLM inference but also enhances response quality by enabling better adaptation to 3D scenes. However, existing methods~\citep{hong20233d, chen2023ll3da, huang2023chat, chen2024grounded} that use learnable 3D scene representations for vision-language tasks typically rely only on spatial coordinates and fail to incorporate semantic relationships between objects - limiting the expressiveness and reasoning capabilities of the model.


In this paper, we introduce 3DGraphLLM, a novel learnable representation of a 3D scene graph designed for use as input to an LLM (see ~\cref{ga}). The representation consists of a list of learnable embeddings for scene objects, where each object is modeled as a local subgraph that includes the object itself and its nearest neighbors. These subgraphs are provided to the LLM as a sequence of triplets \textit{(object1, relation, object2)}. Semantic relations are encoded using features derived from the semantic edges of the scene graph, generated by state-of-the-art methods such as VL-SAT~\citep{wang2023vl}. Our experiments show that incorporating semantic relationships between objects significantly improves the accuracy of LLM responses in 3D vision-language tasks, outperforming baseline methods that use learnable scene representations without semantic context.

\textbf{To summarize}, our contributions are as follows:
\begin{itemize}


\item We introduce 3DGraphLLM, the first method for creating a learnable 3D scene graph representation specifically designed for LLMs. It enables semantic relationships between objects in a scene to be mapped directly into the LLM’s token embedding space.


\item We propose an algorithm that generates a flat sequence of graph embedding tokens by selecting object subgraphs using k-nearest neighbors with Non-Maximum Supression (NMS) and a minimum distance filters between objects. This approach reduces the number of tokens needed to describe the scene, thereby improving inference speed.


\item 3DGraphLLM outperforms the baseline method which does not use semantic relationships on the 3D referred object grounding task, achieving improvements of +7.5\% F1@0.5 on the Multi3DRefer\citep{zhang2023multi3drefer} and +6.4\% Acc@0.5 on ScanRefer~\citep{chen2020scanrefer}  benchmarks. It also improves performance on 3D scene captioning, with a +3.9\% CIDEr@0.5 score on the Scan2Cap~\cite{chen2021scan2cap}  dataset. 3DGraphLLM achieves state-of-the-art results in 3D referred object grounding while requiring up to five times less inference time compared to LVLM-based methods.

\end{itemize}

\section{Related works}

\textbf{3D Language Scene Understanding}. 3D scene understanding is a complex computer vision task that involves identifying the semantic, physical, and functional properties of objects, as well as their mutual relations. One of the goals of 3D scene understanding is to develop methods capable of responding to natural language queries about the scene. The queries may correspond to different visual-language tasks such as 3D referred object grounding~\citep{chen2020scanrefer, zhang2023multi3drefer, miyanishi2024cross3dvg}, question answering~\citep{azuma2022scanqa}, and dense scene captioning~\citep{chen2021scan2cap}. Recent approaches address these queries by reconstructing the scene as a 3D mesh~\citep{peng2023openscene} or point cloud~\citep{zhao20213dvg, chen2022language, zhu20233d}, often enhanced with instance segmentation~\citep{zhu20233d}. 

The emergence of transformer models~\citep{vaswani2017attention} has enabled the development of neural network models that create a learnable representation of a scene for answering various language queries. MultiCLIP~\citep{delitzas2023multi} proposes to align 3D scene representation with text queries and multi-view 2D CLIP~\citep{radford2021learning} embeddings to improve the quality of question answering. 3DVG-Transformer~\citep{zhao20213dvg} and Vil3DRef~\citep{chen2022language} methods introduce modules for modeling spatial relationships between objects to improve the quality of object grounding. 3D-VisTA~\citep{zhu20233d} presents a transformer model for aligning 3D object and text representations, coupled with an unsupervised pre-training scheme to solve various 3D vision-text problems using specialized task-specific heads. However, these approaches face challenges in generalizing to new tasks and domains. In contrast, leveraging large language models (LLMs) for scene understanding enhances generalization capabilities and taps into the extensive knowledge LLMs contain about the physical world~\citep{hong20233d}.

\textbf{Scene Graphs.} The concept of a scene graph was initially developed for 2D images, providing a structured representation of a scene's semantics by incorporating relationships between the semantic elements~\citep{johnson2015image}. In the context of images, scene graphs have proven effective for tasks such as content-based image retrieval~\citep{johnson2015image, pei2023scene}, 2D referring expression comprehension~\citep{yang2019cross, shi2023open, han2024zero}, image caption~\citep{yang2019auto, phueaksri2023approach}, image generation~\citep{johnson2018image, farshad2023scenegenie}.

In 3D scenes, a scene graph is commonly used to address robotics challenges such as planning~\citep{werby2024hierarchical, honerkamp2024language}, object grounding for navigation~\citep{werby2024hierarchical, gu2024conceptgraphs, linok2024beyond, honerkamp2024language} and manipulation~\citep{honerkamp2024language}, as well as scene generation~\citep{zhai2024commonscenes, gao2024graphdreamer}. 
Our approach is part of a class of methods that utilize an implicit representation of the scene graph, such as OVSG~\citep{chang2023context}, which frames the problem of 3D object grounding as subgraph retrieval. 3DGraphQA~\citep{wu20243d} proposes to use the bilinear graph neural network for feature fusion between scene and question graphs for question answering task. FFL-3DOG~\cite{feng2021free} builds a graph based on a text query, which is used to refine the visual graph to select from its vertices the one that best fits the description. However, the application scope of this method is limited to specific tasks such as 3D referred object grounding or question answering. 
In contrast, we propose a more versatile method capable of solving various 3D vision-language tasks.

\textbf{Large Language Models for Scene Understanding.} Large language models (LLMs) offer several advantages for scene understanding, notably enhancing the ability to address complex queries that require common knowledge. LLMs can serve as agents that decompose user queries into elementary tasks, which can then be addressed by other methods~\citep{yang2024llm, yuan2024visual}. Additionally, LLMs can act as an interface for reasoning by processing textual descriptions of the scene as input~\citep{linok2024beyond, gu2024conceptgraphs}. BBQ~\citep{linok2024beyond} and ConceptGraphs~\citep{gu2024conceptgraphs} demonstrate that using a text-based graph representation with an LLM interface significantly improves the quality of object retrieval compared to using CLIP features of objects. HOV-SG~\citep{werby2024hierarchical} constructs a hierarchical graph consisting of objects, rooms, and floors, and demonstrates the effectiveness of such a representation for the task of object grounding given a query containing object location hints. The authors of the MOMA~\citep{honerkamp2024language} method propose using a hierarchical scene graph together with a navigational Voronoi graph as input to LLM to predict a high-level policy for object search for navigation and manipulation. However, using text to describe an object in a scene graph inevitably leads to the loss of some of the information contained in its RGB point cloud. Additionally, in the case of using a text graph, several hundred tokens may be required to describe one object (its semantic class, pose), which will significantly slow down LLM inference in the case of a large number of objects in the scene.

Recent advancements have successfully integrated point cloud data into LLMs by employing pre-trained point cloud encoders and training adapters to align the resulting representations with the LLM embedding space. 3D-LLM~\citep{3dllm} aggregates 3D point cloud features from a sequence of 2D images and then solves the grounding problem as a prediction of a sequence of location tokens added to the LLM dictionary. Chat-Scene~\citep{huang2024chat} generates 2D and 3D features for each object in the scene and introduces learnable object identifier tokens to solve object grounding, dense scene captioning, and question answering problems. LLA3D~\citep{chen2023ll3da} proposes to use a set of trainable fixed-length query tokens obtained by interacting potential visual cues, text cues, and object point cloud features in a transformer model. Grounded 3D-LLM~\citep{chen2024grounded} uses referent tokens to decode object masks in point clouds.  Additionally, research has demonstrated that incorporating spatial information, such as object coordinates~\citep{huang2023chat} or depth maps~\citep{cheng2024spatialrgpt}, enhances the accuracy of responses to user queries.

Despite recent advances, existing methods do not fully leverage the rich semantic information in object relationships. In this paper, we introduce 3DGraphLLM, a method that demonstrates the effectiveness of utilizing semantic relationships between objects to enhance performance across various scene understanding tasks.

\section{Method}

\begin{figure*}[t]
  \centering
  \includegraphics[width=1\linewidth]{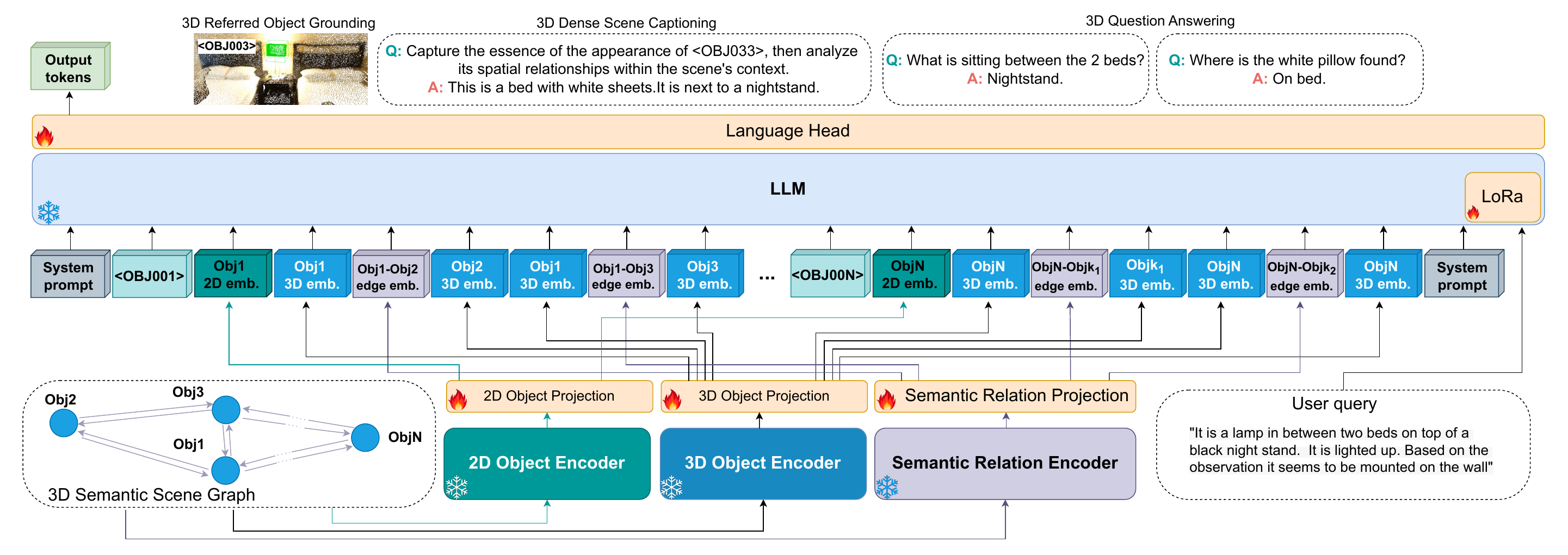}
  \caption{\label{3DGraphLLM-Scheme}
  The overall architecture of our approach. 
  We introduce trainable layers to map the extracted graph node and edge features into the token embedding space of a pre-trained LLM. The scene graph is flattened for input into the LLM, with each object represented by a subgraph of its \textit{k} nearest neighbors. To further adapt the LLM to 3D vision-language tasks, we add new object tokens to the LLM's vocabulary alongside with objects' 2D features and fine-tune the LLM using LoRa.
  }
\end{figure*}

Our approach uses a set of point clouds of scene objects as input. The objects' point clouds can be obtained either from ground-truth annotations or through state-of-the-art point cloud instance segmentation methods. These point clouds are used to extract scene graph features (see ~\cref{sec:architecture}). A scene graph consists of nodes representing the objects and edges corresponding to semantic relationships between them.  To convert the scene graph into a token sequence, we represent each object by an identifier, its 2D object feature, and a subgraph comprising the object's \textit{k} nearest neighbors. The relationships between an object and its neighbors are encoded as triplets $(object_i, relation_{ij}, object_j)$. The scheme of the 3DGraphLLM approach is shown in ~\cref{3DGraphLLM-Scheme}. For more details on the scene graph representation, refer to ~\cref{sec:graph-representation}. Our training process is two-stage. First, we pre-train the model on a dataset for various 3D scene understanding tasks using ground-truth instance segmentation. Next, we fine-tune 3DGraphLLM with predicted instance segmentation of scene point clouds, considering a scenario where ground-truth segmentation is unavailable (see ~\cref{sec:training}).

\subsection{Model Architecture}
\label{sec:architecture}

The model architecture includes pre-trained encoders for 2D images, 3D point clouds, and point clouds semantic relationships, alongside a pre-trained LLM.  We train projection layers to map the extracted object features and their relationships into the LLM's token embedding space. Following the approach of Chat-Scene~\citep{huang2024chat}, we introduce additional object identifier tokens $\lbrace <\verb+OBJ+i> \rbrace_{i=1}^{n}$ into the LLM's vocabulary. Here and throughout, we use $n$ to denote the number of objects in the scene. These learned identifiers, with the features from object subgraphs composed of nearest neighbors for each object, are used to create a flat representation of the scene graph, which is then fed into the LLM. 

\textbf{Object Proposals.}
We use point clouds of objects in the scene as vertices in the scene graph $\mathit{G}$. In our experiments, we evaluate 3DGraphLLM in various modes, including ground-truth scene segmentation and instance segmentation using state-of-the-art neural network methods like Mask3D~\citep{schult2023mask3d} and OneFormer3D~\citep{kolodiazhnyi2024oneformer3d}. Thus, the set $V$ of vertices of the graph consists of $n$ point clouds $\lbrace P_{i} \rbrace_{i=1}^{n}$, where $ P_{i} \in \mathbb{R}^{m_{i} \times 6} $. Here, $m_i$ is the number of points in the $i$-th object proposal of instance segmentation of scene point cloud, and $6$ dimensions of each point correspond to its 3D coordinates and RGB color.

\textbf{Object Identifiers.}
Following the approach in Chat-Scene, we add a set of learnable identifier tokens  $\lbrace <\verb+OBJ+i> \rbrace_{i=1}^{n}$ to the LLM's vocabulary for object identification. These tokens allow the model to identify objects in the scene by simply predicting the corresponding object identifier token. In our experiments, we assume a maximum of 200 objects per scene.

\textbf{2D Object Encoder.} The results of Chat-Scene demonstrate that adding aggregated 2D DINOv2\citep{oquab2023dinov2} features increases the LLM performance on 3D vision-language tasks. Therefore, we add DINOv2 $ Z_{i}^{2d} \in \mathbb{R}^{1 \times 1024} $ features as an additional token describing the object subgraph. DINOv2 object features are obtained by aggregating features from the masked multi-view images where masks come from the projection of the object's 3D point cloud. 

\textbf{3D Object Encoder.} 
We extract vertex features using a pre-trained Uni3D~\citep{zhou2023uni3d} encoder, which generates point cloud features aligned with their textual descriptions. Since this model is pre-trained on a large dataset, it enables us to produce high-quality graph vertex embeddings across various data domains. For each object point cloud $ P_{i} $, we extract Uni3D feature $ Z_{i}^{v_p}  \in \mathbb{R}^{1 \times 1024} $. 

\textbf{Edge Feature Encoder.}
One challenge in generating features for semantic relationships between objects is that most methods for 3D semantic scene graph generation are trained on 3RScan scenes~\citep{wald2019rio}, while visual grounding tasks are typically tested on ScanNet scenes~\citep{dai2017scannet}. Although both datasets belong to the indoor scene domain, existing methods struggle with performance in cross-domain testing, resulting in a drop in accuracy for the grounding task~\citep{miyanishi2024cross3dvg}.

To extract semantic relationships between objects, we use VL-SAT~\citep{wang2023vl}, a method for generating 3D semantic scene graphs from point clouds. One of its key advantages is that it only requires 3D point cloud coordinates as input during prediction while leveraging knowledge transfer from the pre-trained CLIP model~\citep{radford2021learning}. This allows the method to perform well when applied to new scene domains~\citep{wang2023vl}, as confirmed by our experiments (see ~\cref{sec:ablation}). For each pair of point clouds $ P_{i} $ and $ P_{j} $, we generate a latent feature representing their relationship $ Z_{ij}^e \in \mathbb{R}^{1 \times 512}$, which corresponds to VL-SAT graph neural network feature before the classification head assigning semantic categories to the graph edges. While VL-SAT predicts a fixed set of relationships between objects, these relationships are not mutually exclusive (e.g., "larger" and "close"). Therefore, we use latent features to capture possible combinations of these semantic relationships.

\textbf{2D/3D object, and semantic relation projection.} 
To adapt the extracted features for the language model, we use three trainable projection modules: the 2D Object Projection $f_{2d}(\cdot)$, which maps the 2D image features of objects, the 3D Object Projection $f_v(\cdot)$, which maps the point cloud features of objects, and the Semantic Relation Projection $f_e(\cdot)$, which maps the features of semantic relationships between objects. Therefore, for the $i$-th object, the 2D and 3D object features are projected to token embeddings $F_{i}^{v}$ and $F_{i}^{2d}$, respectively. For the pair of $i$-th and $j$-th objects, the semantic relation feature is projected to token embedding $F_{ij}^{e}$:

\begin{equation}
F_{i}^{2d} = f_v( Z_{i}^{2d}), F_{i}^{v} = f_v( Z_{i}^v), F_{ij}^{e} = f_e(Z_{ij}^e).
\end{equation}

\subsection{Flat Graph Representation}
\label{sec:graph-representation}

The scene graph is a complete graph since we can generate connections between all pairs of objects. Such a graph contains $n\cdot (n-1)$ edges between objects, and using the complete graph as a sequence for the LLM would significantly increase the sequence length. Intuitively, the most relevant relationships for answering user questions are those between an object and its nearest neighbors. Therefore, for each object, we consider a subgraph of its \textit{k} nearest neighbors. The relationships between objects are encoded using features extracted from point clouds $ \lbrace F_{i}^{v} \rbrace_{i=1}^{n}$ and semantic relations features $ \lbrace F_{ij}^{e}, i \in \lbrace 1, ..., n \rbrace, j \in \lbrace 1, ..., n \rbrace \rbrace$, represented as a triplet $(F_{i}^{v}, F_{ij}^{e}, F_{j}^{v})$.

When using the complete scene graph the number of tokens required to describe the scene is $2\cdot n + 3n \cdot (n - 1)$. For $100$ objects, which matches the number of object proposals in the Mask3D~\citep{schult2023mask3d} instance segmentation, this totals $29 900$ tokens. By using a $k$-nearest neighbor subgraph, we reduce the token count to $2\cdot n + 3n \cdot k$. As shown in ~\cref{sec:ablation} (see ~\cref{nn-ablation}) and Supplementary Materials, setting $k=2$ improves accuracy in 3D visual-language tasks while reducing the number of tokens needed to describe a scene with $100$ objects to $800$. We analyze how the number of objects affects inference speed and GPU memory usage in Supplementary Materials.

\textbf{Prompt template.} 
We integrate the scene description as a sequence of object subgraphs into the prompt for LLM similar to the integration of the list of object embeddings in the Chat-Scene method~\citep{huang2024chat}. An example of a prompt for LLM containing a system prompt, a scene description in the form of an object identifier, a 2D object feature and an object subgraph, a user request, and an LLM assistant response is given in ~\cref{tab:prompt}. The sequence describing an object $i$ starts with its identification token \verb+<OBJi>+ and 2D object feature $F_{i}^{2d}$. Then there are $k$ triplets $ \lbrace (F_{i}^{v}, F_{ij_k}^{e}, F_{j_k}^{v}) \rbrace_{j_k=1}^{k}$ describing the relationship between the object and its $k$ nearest neighbors.

\begin{table}[t]
\scriptsize
\centering

\begin{tabular}{|rp{6.5cm}|}
\hline
System: & A chat between a curious user and an artificial intelligence assistant. \\
& The assistant gives helpful, detailed, and polite answers to the user’s questions. The conversation centers around an indoor scene:\verb+[<OBJ001>+ $F_{1}^{2d}, F_{1}^{v}, F_{12}^{e}, F_{2}^{v} F_{1}^{v}, F_{14}^{e}, F_{4}^{v}...$\verb+<OBJN>+ $F_{N}^{2d}, F_{N}^{v}, F_{Nk_1}^{e}, F_{k_1}^{v} F_{N}^{v}, F_{Nk_2}^{e}, F_{k_2}^{v}$]\\ 
User: &  According to the given description, \textit{there are brown wooden cabinets}, \textit{placed on the side of the kitchen},  please provide the ID of the object  that closely matches this description. \\
Assistant:  & \verb+<OBJ001>.+\\
\hline
\end{tabular}
\caption{
 Example of prompt for the language model containing scene graph.
}
\label{tab:prompt}
\end{table}

\subsection{Training Strategy}
\label{sec:training}

Following the strategy used in Chat-Scene\citep{huang2024chat}, we implement a training approach that involves simultaneously training the projection layers and the language model. We also conduct joint training for various tasks, including visual grounding (ScanRefer~\citep{chen2020scanrefer}, Multi3DRefer~\citep{zhang2023multi3drefer}, RioRefer~\citep{miyanishi2024cross3dvg}), 3D scene description (Scan2Cap~\citep{chen2021scan2cap}, Nr3D~\citep{achlioptas2020referit3d}, RioRefer~\citep{miyanishi2024cross3dvg}), and 3D visual question answering (ScanQA~\citep{azuma2022scanqa}, SQA3D~\citep{ma2022sqa3d}, 3RQA~\citep{huang2023embodied}). This adaptation of the tasks is designed for user-assistant interactions, as proposed by the authors of Chat-Scene.
During training, we aim to optimize the trainable parameters $\theta$ of both the language model and the projection layers to minimize the negative log-likelihood of the target response $s^{\text{res}}$ compared to the response predicted by the model. 
We use the following loss function:

\begin{equation}
L(\theta) = - \sum_{i=1}^{\ell}\log{P(s^{\text{res}}_{i}|s^{\text{res}}_{[1,...,i-1]},s^{\text{prefix}})},
\end{equation}

where $\ell$ is the length of the token sequence in the LLM response, $s^{\text{res}}_{[1,...,i-1]}$ is the sequence generated up to the $i$-th token, $s^{\text{prefix}}$ is the input prefix sequence containing system and user prompts.
The trainable parameters $\theta$ include the parameters of 2D/3D Object Projections and Semantic Relation Projection Layers, added object identifier token embeddings, and the language model.

We use the semantic relationships encoder~\citep{wang2023vl} pre-trained using ground-truth (GT) point cloud scene segmentation data. Since the predicted point cloud segmentation typically contains more noise than the GT segmentation, we anticipate that the edge features derived from the GT segmentation will be of higher quality than those from the neural network instance segmentation. 
To address this problem, we employ a two-stage training strategy for 3DGraphLLM. First, we pre-train the projection layers and the language model on the GT instance segmentation data to achieve effective projections of the semantic embeddings of relations and objects into the language model's embedding space. Then, we fine-tune 3DGraphLLM using the noisy data from the neural network segmentation. ~\cref{sec:ablation} presents the experimental results, demonstrating the effectiveness of two-stage training and comparing different pre-training datasets.

\section{Experiments}
\begin{table*}[t]
  \tiny
  \centering
   \begin{tabular}{llcccp{0.6cm}p{0.6cm}p{0.6cm}p{0.6cm}llp{0.6cm}p{0.6cm}p{0.6cm}p{0.6cm}}
    \hline
    \multirow{2}{*}{} &
      \multicolumn{1}{c}{} &
      \multicolumn{1}{c}{} &
      \multicolumn{1}{c}{} &
      \multicolumn{1}{c}{} &
      \multicolumn{2}{c}{ScanRefer} &
      \multicolumn{2}{c}{Multi3DRefer} &
      \multicolumn{2}{c}{Scan2Cap} &
      \multicolumn{2}{c}{ScanQA} &
      \multicolumn{1}{c}{Sqa3D} \\
    & Methods & 2D features & 3D features & LLM & A@0.25\textuparrow & A@0.5\textuparrow & F1@0.25\textuparrow 
    & F1@0.5\textuparrow & C@0.5\textuparrow & B-4@0.5\textuparrow & C\textuparrow & B-4\textuparrow  &  EM\textuparrow   \\
    \hline
    \multirow{9}{*}{\textit{\rotatebox[origin=c]{90}{\textit{Expert models}}}} & ScanRefer~\citep{chen2020scanrefer}  & \cmark & \cmark & \xmark & 37.3 & 24.3 & - & - &  - & - & - & - &  - \\
    & MVT~\citep{huang2022multi}  & \cmark & \cmark & \xmark & 40.8 & 33.3 & - & - & - & - & - & - & - \\
    & 3DVG-Trans~\citep{zhao20213dvg}  & \cmark & \cmark & \xmark & 45.9 & 34.5 &  - & - & - & - & - & - & - \\
    & ViL3DRel~\citep{chen2022language}  & \xmark & \cmark & \xmark & 47.9 & 37.7 & - & - & - & - &  - & - & - \\
    & M3DRef-CLIP~\citep{zhang2023multi3drefer}  & \cmark & \cmark &  \xmark & 51.9 & 44.7 & 42.8 & 38.4 & - & - & - & - & - \\
    & Scan2Cap~\citep{chen2021scan2cap}  & \cmark & \cmark &  \xmark & - & - & - & - & 35.2 & 22.4 & - & - & -  \\
    & ScanQA~\citep{azuma2022scanqa} & \cmark & \cmark &  \xmark & - & - & - & - & - & - & 64.9 & 10.1 & -  \\
    & Sqa3D~\citep{ma2022sqa3d}  & \xmark & \cmark &  \xmark & - & - & - & - & - & - & - & - & 47.2  \\
    & 3D-VisTA~\citep{zhu20233d}  & \xmark & \cmark &  \xmark & 50.6 & 45.8 & - & - & 66.9 & 34.0 & 72.9 & 13.1 & 48.5  \\
    & BUTD-DETR~\citep{jain2022bottom} & \xmark & \cmark &  \xmark & 52.2 & 39.8 & - & - & - & - & - & - & - \\
    & PQ3D~\citep{zhu2025unifying} & \cmark & \cmark & \xmark &  - & 51.2 & - & 50.1 & 80.3 & 36.0 & 87.8 & - & 47.1  \\
    \hline
    \multirow{14}{*}{\rotatebox[origin=c]{90}{\textit{LLM-based models}}} & ZSVG3D~\citep{yuan2024visual}   & \cmark & \cmark & GPT4 & 36.4 & 32.7 &   & - &  - & - & - & - &  - \\
    & 3D-LLM~\citep{3dllm}  & \cmark & \cmark & Flamingo & 21.2 & - & - & - & - &  - & 59.2 & 7.2 & -  \\
    & 3D-LLM~\citep{3dllm} & \xmark & \cmark & BLIP2-flant5 & 30.3 & - & -  & - & - &  - & 69.4 &  12.0 & -  \\
    & Chat-3D v2~\citep{huang2023chat}  & \xmark & \cmark & Vicuna-7B-v0 & 35.9 & 30.4 & - & - & - &  - & 77.1 & 7.3 & -  \\
    & Scene-LLM~\citep{fu2024scene}  & \cmark & \cmark & Llama-2-7B & - & - & -  & - & - &  - &  80.0 & 12.0 & 54.2  \\
    & LEO~\citep{huang2023embodied}  & \xmark & \cmark & Vicuna-7B-v1.1 & - & - & -  & - & 72.4 & 38.2 & \textbf{101.4} & 13.2 &  50.0 \\
    & LL3DA~\citep{chen2023ll3da}  & \xmark & \cmark & OPT-1.3B & - & - & -  & - & 65.2 &  36.8  & 76.8 & 13.5 & - \\
    & Grounded 3D-LLM~\citep{chen2024grounded}  & \xmark & \cmark & Tiny-Vicuna-1B & 47.9 & 44.1 & 45.2 & 40.6 & 70.6 &  35.5 & 72.7 & 13.4 & -  \\
    
    & Robin3D~\citep{kang2025robin3dimproving3dlarge} & \cmark & \cmark & Vicuna-7B-v1.5 & 60.8 & 55.1 & \textbf{64.9} & 59.7 & \textbf{87.2} & \underline{38.4} & - & - & 56.0  \\
    & GPT4Scene-HD~\citep{qi2025gpt4scene}  & \cmark & \cmark & Qwen2-VL-7B & 50.9 & 46.4 & 53.7 & 50.0 & 74.4 & 37.9 & 89.9 & \textbf{15.9} & \underline{57.2} \\
    & GPT4Scene-HDM~\citep{qi2025gpt4scene} & \cmark & \cmark  & Qwen2-VL-7B & \textbf{62.6} & \textbf{57.0} & 64.5 & \underline{59.8} & \underline{86.3} & \textbf{40.6} & \underline{96.3} & \underline{15.5} & \textbf{59.4} \\
    \cline{2-14}
    & Chat-Scene~\citep{huang2024chat} (baseline) & \cmark & \cmark & Vicuna-7B-v1.5 & 55.5 & 50.2 & 57.1 & 52.4 & 77.1 & 36.3 & 87.7 & 14.3 & 54.6 \\
    & 3DGraphLLM  (ours) & \cmark & \cmark  & Vicuna-7B-v1.5 & 58.6 & 53.0 & 61.9 & 57.3 & 79.2 & 34.7 & 91.2 & 13.7 & 55.1 \\
    & 3DGraphLLM (ours) & \cmark & 
    \cmark &  LLAMA3-8B-Instruct & \underline{62.4} & \underline{56.6} & \underline{64.7} & \textbf{59.9} & 81.0 & 36.5 & 88.8 & \textbf{15.9} & 55.9 \\
  \hline
  \end{tabular}
  \caption{
    Performance comparison of 3DGraphLLM with state-of-the-art approaches for 3D vision-language tasks. "Expert models" use specialized heads to deal with different 3D vision-language tasks. Our approach falls into the category of "LLM-based models" that consider different tasks as different user queries to a generative model. C denotes the CIDEr metric.
  }
 \label{tab:scannet}
\end{table*}
\textbf{Datasets.}
For pretraining 3DGraphLLM using GT instance segmentation, we employ a combined 3D Vision-Language dataset for ScanNet~\citep{dai2017scannet} and 3RScan~\citep{wald2019rio} scenes. For ScanNet scenes, we utilize data from five 3D vision-language benchmarks: visual grounding tasks (ScanRefer~\citep{chen2020scanrefer}, Multi3DRefer~\citep{zhang2023multi3drefer}), scene description (Scan2Cap~\citep{chen2021scan2cap}), and 3D visual question answering (ScanQA~\citep{azuma2022scanqa}, SQA3D~\citep{ma2022sqa3d}). Each of these datasets follows a standard split into training and validation sets, corresponding to 1201 training scans and 312 validation scans from ScanNet. For 3RScan scenes, we use data from the RioRefer dataset~\citep{miyanishi2024cross3dvg} for object grounding, and the 3RQA dataset~\citep{huang2023embodied} for question answering. For 3RScan data, we follow the standard train/validation scan split and use the scans present in the RioRefer dataset for training, resulting in 1175 training scans and 157 validation scans. To augment the data for the scene description task, we use data from the RioRefer~\citep{miyanishi2024cross3dvg} and Nr3D~\citep{achlioptas2020referit3d} datasets, taking object grounding queries provided in these datasets as reference descriptions of objects in the scene.
To assess 3DGraphLLM performance under realistic conditions, we perform fine-tuning on predicted instance segmentation using 3D vision-language benchmarks for ScanNet scenes: ScanRefer, Multi3DRefer, Scan2Cap, ScanQA, and SQA3D.

\textbf{Implementation details.}
The projection layers for 2D/3D object features and their semantic relations are three-layer MLPs. In our experiments, we use LLAMA3-8B-Instruct~\citep{llama3modelcard}, a state-of-the-art large language model, as well as Vicuna-1.5-7B~\citep{zheng2023judging} for ablation. For fine-tuning the language model, we apply LoRA~\citep{hu2021lora} with a rank of 16. We use a batch size of 8 and train 3DGraphLLM for $3$ epochs with an initial learning rate of $5\cdot 10^{-6}$, following a cosine annealing schedule. Training is performed on a server equipped with $4$ NVIDIA A100 GPUs, and the entire training process takes approximately $24$ hours. In our experiments, we select $k=2$ nearest neighbors to construct object subgraphs and, in the case of using Mask3D~\citep{schult2023mask3d} instance scene point cloud segmentation, we use a NMS filter and a filter that ensures a minimum distance between nearest neighbors of 1 cm (see ~\cref{sec:ablation}).

\begin{table}[h]
\tiny
\centering
\begin{tabular}{llll}
    \hline
    \multirow{2}{*}{} &
      \multicolumn{1}{l}{} &
      \multicolumn{2}{c}{Method} \\
     & Dataset & 3DGraphLLM & GPT4Scene  \\
    \hline
    Input token number per scene & & \textbf{800} & 10400 \\
    \hline
    \multirow{5}{*}{Inference speed, sec} & ScanRefer & \textbf{0.4} & 1.9 \\
    & Multi3DRefer & \textbf{0.5} & 2.0 \\
    & Scan2Cap & \textbf{0.9} & 2.2 \\
    & ScanQA & \textbf{0.4} & 1.9 \\
    & SQA3D & \textbf{0.4} & 1.7 \\
  \hline
\end{tabular}
\caption{Input tokens and inference speed comparison (Mask3D instance segmentation).}
\label{tab:speed}
\end{table}

\textbf{Evaluation metrics.} 
For the visual grounding task on the ScanRefer~\citep{chen2020scanrefer}dataset, we use the standard metrics Acc@0.25 and Acc@0.5. A prediction is considered a true positive if the intersection-over-union (IoU) between the predicted object's 3D bounding box and the ground truth exceeds the thresholds of 0.25 and 0.5, respectively. The Multi3DRefer~\citep{zhang2023multi3drefer} dataset contains queries that may refer to multiple objects. Therefore, we use the benchmark-standard F1 score at IoU thresholds of 0.25 and 0.5. We assess the quality of object descriptions using the Scan2Cap~\citep{chen2021scan2cap} benchmark metrics CIDEr@0.5 and BLEU-4@0.5. For the visual question answering task, we follow the validation strategy from Chat-Scene\citep{huang2024chat}, applying CIDEr~\citep{vedantam2015cider} and BLEU-4~\citep{papineni2002bleu} metrics for ScanQA~\citep{azuma2022scanqa}, and exact match accuracy (EM) for SQA3D~\citep{ma2022sqa3d}.

\begin{figure*}[t]
  \centering
  \includegraphics[width=0.6\linewidth]{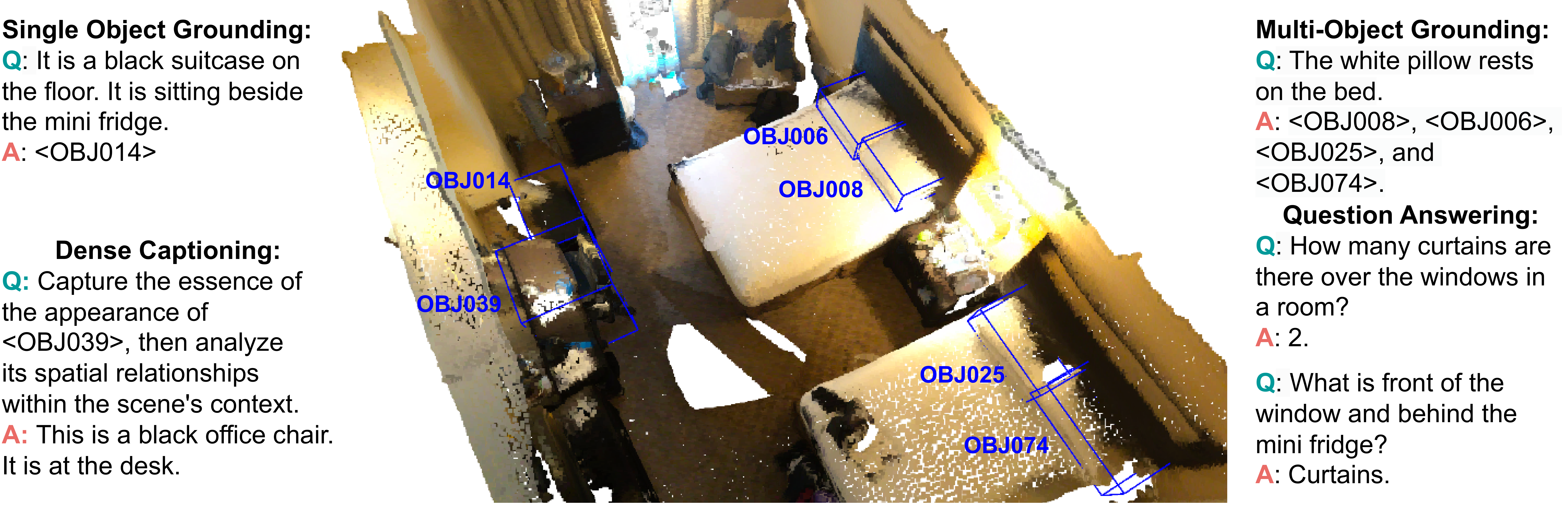}
  \caption{\label{fig:3DGraphLLM-Qualitative}
  Qualitative examples of 3DGraphLLM performance on object grounding, dense captioning, and question answering tasks. We provide a visualization of the RGB point cloud along with blue objects bounding boxes.
 }
\end{figure*}
\begin{table*}[t]
\tiny
\centering
\begin{tabular}{llllllllllll}
    \hline
      \multicolumn{1}{c}{} &
      \multicolumn{1}{c}{} &
      \multicolumn{1}{p{0.5cm}}{} &
      \multicolumn{1}{p{0.9cm}}{Number} &
      \multicolumn{1}{p{1.2cm}}{Training scenes} &
      \multicolumn{1}{c}{ScanRefer} &
      \multicolumn{1}{c}{Multi3DRefer} &
      \multicolumn{2}{c}{Scan2Cap} &
      \multicolumn{2}{c}{ScanQA} &
      \multicolumn{1}{c}{Sqa3D} \\
     Methods & LLM & Pre-train & of edges & & Acc@0.5\textuparrow & F1@0.5\textuparrow & C@0.5\textuparrow & B-4@0.5\textuparrow  & C\textuparrow & B-4\textuparrow & EM\textuparrow \\
    \hline

    3DGraphLLM-0 & Vicuna1.5-7B  & \xmark & 0 & ScanNet & 50.2 & 52.4 & 77.1 & \underline{36.3} & 87.7 & \underline{14.3} & 54.6  \\
    3DGraphLLM-2 & Vicuna1.5-7B & \xmark & 2 & ScanNet & 50.1 & 52.7 & \textbf{80.4} & \textbf{36.9} & \textbf{92.2} & \textbf{15.5} & \underline{54.7} \\
    3DGraphLLM-2 & Vicuna1.5-7B & \cmark & 2 & ScanNet+3RScan & \textbf{53.1} & \textbf{57.3} & \underline{79.2} & 34.7 & \underline{91.2} & 13.7 & \textbf{55.1} \\
    \hline
    3DGraphLLM-0 & LLAMA3-8B-Instruct & \xmark & 0 & ScanNet & 52.0 & 55.1 & 80.0 & 37.5 & 84.0 & \underline{15.8} & 53.8  \\
    3DGraphLLM-2 & LLAMA3-8B-Instruct & \xmark & 2 & ScanNet & 54.3 & 57.3 & \textbf{85.6} & \textbf{39.6} & \underline{87.4} & 14.9 & 54.5  \\
    3DGraphLLM-2 & LLAMA3-8B-Instruct & \cmark & 2 & ScanNet & \underline{56.2} & \underline{58.7} & \underline{82.9} &  \underline{37.3} & 85.4 & \underline{15.1} & \underline{55.6} \\
    3DGraphLLM-2 & LLAMA3-8B-Instruct & \cmark & 2 & ScanNet+3RScan & \textbf{56.6} & \textbf{59.9} & 81.0 & 36.5 & \textbf{88.8} & \textbf{15.9 }& \textbf{55.9} \\
  \hline
\end{tabular}
\caption{Ablation study on semantic edges role and training pipeline. C denotes the CIDEr metric.}
\label{tab:semantic-edges-gt}
\end{table*}

\subsection{Experimental Results}
\textbf{Comparison with state-of-the-art approaches.}
As shown in ~\cref{tab:scannet}, our method significantly outperforms the baseline approach Chat-Scene~\cite{huang2024chat} on the two ScanNet 3D referred object grounding benchmarks, ScanRefer~\citep{chen2020scanrefer} and Multi3DRefer~\citep{zhang2023multi3drefer}, as well as on the scene captioning benchmark Scan2Cap~\citep{chen2021scan2cap} and the question answering benchmarks ScanQA~\citep{azuma2022scanqa} and SQA3D~\citep{ma2022sqa3d}. These results highlight the effectiveness of a learnable graph-based scene representation 3D vision-language tasks. It's worth noting that the performance of our method surpasses state-of-the-art specialized models with separate heads for different language tasks, such as 3D-VisTA~\citep{zhu20233d}, PQ3D~\citep{zhu2025unifying}, and M3DRef-CLIP~\citep{zhang2023multi3drefer}.

 Notably, 3DGraphLLM demonstrates state-of-the-art quality for the 3D referred object grounding task for LLM-based methods. In particular, our 3DGraphLLM with LLAMA3-8B as the base LLM outperforms Robin3D~\citep{kang2025robin3dimproving3dlarge} on ScanRefer benchmark showing comparable quality on Multi3DRefer and SQA3D benchmarks. Robin3D is trained on 1M instruction-following data that are not publicly available, while our approach uses only 370K instruction-following data. Our experiments in ~\cref{tab:semantic-edges-gt} highlight the importance of training data for 3DGraphLLM, suggesting that incorporating more data for fine-tuning could further improve its performance. 3DGraphLLM achieves results comparable to the state-of-the-art method GPT4Scene-HDM~\citep{qi2025gpt4scene}, showing the importance of semantic relations for this task. At the same time, 3DGraphLLM uses fewer tokens to describe the scene (see~\cref{tab:speed}), allowing up to five times faster inference for object-grounding tasks.

\textbf{Qualitative results.} ~\cref{fig:3DGraphLLM-Qualitative} shows the qualitative results of 3DGraphLLM using Mask3D~\citep{schult2023mask3d} instance scene segmentation. 
3DGraphLLM efficiently uses spatial cues for solving 3D Vision-Language tasks. For example, 3DGraphLLM distinguishes the black suitcase next to the refrigerator, despite there being another suitcase farther away from the refrigerator in the scene. In Supplementary Materials we provide more examples of 3DGraphLLM performance.

\subsection{Ablation Studies}
\label{sec:ablation}

\textbf{Role of Semantic Relations.} To isolate the impact of using a scene graph representation, we conduct an experiment with different LLMs and training pipelines using Mask3D~\citep{schult2023mask3d} instance segmentation. We train a version of 3DGraphLLM (3DGraphLLM-0) where the scene is represented as a sequence of object identifiers and features extracted by the 2D Object Encoder and the 3D Object Encoder, following the same training pipeline as 3DGraphLLM (3DGraphLLM-2) with two nearest neighbors. The 3DGraphLLM version with zero nearest neighbors serves as a baseline, equivalent to the Chat-Scene approach, which uses the same LLM as 3DGraphLLM-2.
As shown in ~\cref{tab:semantic-edges-gt}, incorporating a scene graph representation significantly improves the performance of the LLMs across all three 3D Vision-Language tasks: visual grounding, scene description, and question answering. 
However, the effect is more noticeable for the more recent LLAMA3-8B-Instruct. 

\textbf{Training pipeline.} The pre-training on GT instance segmentation data improves the quality of the 3D referred object grounding for LLAMA3-8B-Instruct and Vicuna-1.5-7B. For LLM Vicuna-1.5-7B, pre-training increases the scene captioning quality. For LLAMA3-8B-Instruct, pre-training improves the question answering on the SQA3D dataset. We compare two pre-training datasets for 3DGraphLLM using LLAMA3-8B-Instruct. The first contains only 3D Vision-Language data from ScanNet, while the second includes data from both ScanNet and 3RScan. ~\cref{tab:semantic-edges-gt} shows that incorporating 3RScan data further enhances object grounding and question answering performance.
The most interpretable metrics for the role of semantic edges are the accuracy metrics in the 3D referred object grounding task, so we keep this pre-training as part of the 3DGraphLLM training pipeline.


It is worth noting that the n-gram-based evaluation metrics used in scene captioning and question answering benchmarks are not adequate for assessing the quality of LLM-generated responses because they fail to capture the flexibility and richness of LLM outputs. This effect is particularly noticeable in the scene captioning task, where CIDEr@0.5 and BLEU-4@0.5 penalize 3DGraphLLM if the model incorporates visual and spatial cues that are missing from the reference descriptions. For example, in the scene shown in ~\cref{fig:3DGraphLLM-Qualitative}, 3DGraphLLM describes a toilet as: \textit{"This is a white toilet. It is to the right of the shower curtain."} This is a correct description of the object, yet the reference captions use different wording and spatial cues, causing CIDEr@0.5 to assign a score of 0.0 to this description. See Supplementary Materials for a more detailed illustration of this effect.

\begin{table}[t]
\tiny
\centering
\begin{tabular}{p{1.3cm}p{1.7cm}p{0.6cm}p{0.9cm}p{0.15cm}p{0.15cm}}
    \hline
      \multicolumn{1}{c}{} &
      \multicolumn{1}{l}{Instance  } &
      \multicolumn{1}{l}{Number } &
      \multicolumn{1}{l}{Minimal } &
      \multicolumn{1}{c}{ScanRefer} &
      \multicolumn{1}{c}{Multi3DRef}\\
     Methods & segmentation &  of edges &  distance, cm & Acc@0.5\textuparrow & F1@0.5\textuparrow \\
    \hline
    3DGraphLLM-0 & GT & 0 & - & 61.5 & 64.4 \\
    3DGraphLLM-2 & GT & 2 & 0 & \textbf{66.9} & \textbf{69.9}  \\
    \hline
    3DGraphLLM-0 & Mask3D & 0 & - & 52.0 & 55.1 \\
    3DGraphLLM-2 & Mask3D & 2 & 0 & 55.6 & 58.2 \\
    3DGraphLLM-2 & Mask3D (+ NMS) & 2 & 0 & \underline{55.7} & \underline{58.6} \\
    3DGraphLLM-2 & Mask3D (+ NMS) & 2 & 1 & \textbf{56.2} & \textbf{58.7}  \\
    \hline
    3DGraphLLM-0 & OneFormer3D & 0 & - & 50.0 & 52.8 \\
    3DGraphLLM-2 & OneFormer3D & 2 & 0 & \underline{52.8} & \underline{55.8}  \\
    3DGraphLLM-2 & OneFormer3D (+NMS) & 2 & 1  & \textbf{54.6} & \textbf{57.2} \\
  \hline
\end{tabular}
\caption{Ablation study on semantic edges role depending on quality of instance segmentation. }
\label{tab:semantic-edges-inst-segm}
\end{table}

\begin{table}[h]
\tiny
\centering
\begin{tabular}{p{1.3cm}p{1cm}p{0.8cm}p{0.6cm}p{0.15cm}p{0.15cm}}
    \hline
      \multicolumn{1}{c}{} &
      \multicolumn{1}{l}{Instance} &
      \multicolumn{1}{l}{Relations } &
      \multicolumn{1}{c}{Number } &
      \multicolumn{1}{c}{ScanRefer} &
      \multicolumn{1}{c}{Multi3DRef}\\
     Methods & segmentation & as triplets & of edges & Acc@0.5\textuparrow & F1@0.5\textuparrow \\
    \hline
    3DGraphLLM-0 & Mask3D & \xmark & 0 & 52.0 & 55.1 \\
    3DGraphLLM-2 & Mask3D & \xmark & 2 & \underline{54.2} & \underline{56.3} \\
    3DGraphLLM-2 & Mask3D & \cmark & 2 & \textbf{54.3} & \textbf{57.3} \\

  \hline
\end{tabular}
\caption{Ablation study on subgraph representation.}
\label{tab:relations-tokens}
\end{table}

\textbf{Quality of instance segmentation.} 
We evaluate how the quality of scene segmentation into objects impacts the performance of 3DGraphLLM. For these experiments, we use the full training pipeline with a pre-training phase on GT instance segmentation on ScanNet data.
As shown in ~\cref{tab:semantic-edges-inst-segm}, even with noisy neural network segmentation, representing the scene as a graph with semantic relationships is still more effective than using a simple list of objects. 
We conduct experiments with different object proposal methods, including OneFormer3D~\citep{kolodiazhnyi2024oneformer3d} and Mask3D~\citep{schult2023mask3d}, but we found that Mask3D segmentation shows better results for our tasks.
Therefore, in subsequent experiments, we use the Mask3D method to maintain consistency with the baseline Chat-Scene approach.

\begin{figure}
  \begin{center}
  \includegraphics[width=1\linewidth]{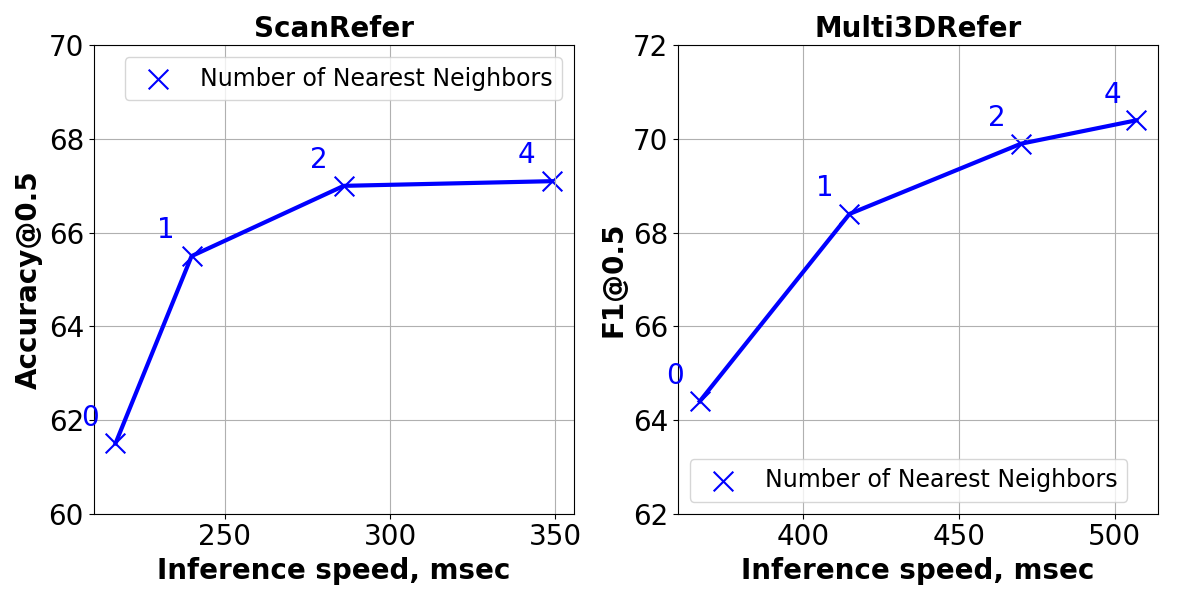}
  \end{center}
  \caption{\label{nn-ablation}
  Dependence of inference speed and visual grounding quality on the number of nearest neighbors in the object subgraph. This experiment utilizes the GT instance segmentation.
  }
\end{figure}



The analysis of objects selected as nearest neighbors reveals a high number of duplicate objects among the chosen neighbors. To address this issue, we propose two filters. First, we add an NMS filter to remove duplicates between the potential neighbors for an object, using a threshold of $IoU=0.99$. Second, we introduce a minimum distance filter of 1 cm to the nearest neighbor to prevent selecting duplicates of the original object as its neighbors.

Adding the NMS filter improves the performance of the visual grounding task when using Mask3D instance segmentation (see ~\cref{tab:semantic-edges-inst-segm}). The additional minimum distance filter further enhances visual grounding quality. The combination of filters is also effective for OneFormer3D~\citep{kolodiazhnyi2024oneformer3d} scene instance segmentation, as shown in ~\cref{tab:semantic-edges-inst-segm}.


\textbf{Number of nearest neighbors.} 
We examine how the number of nearest neighbors affects the quality of visual grounding and the speed of model inference, as adding more connections increases the number of tokens used to describe each object. 
This experiment was performed using ground-truth scene segmentation, as this setup provides the highest quality embeddings for semantic relations between objects. We vary the number of nearest neighbors in powers of two, capping it at 4 due to GPU memory constraints during training. As shown in ~\cref{nn-ablation}, increasing the number of nearest neighbors enhances visual grounding quality with a slight increase in inference time.


\textbf{Subgraph representation.} 
In our work, we use an object-centric graph representation, where relationships between objects are represented as triplets  $ \lbrace F_{N}^{v}, F_{Nk_1}^{e}, F_{k_1}^{v} \rbrace$.
We conduct an experiment in which we remove duplicate vertex tokens from the subgraph-based object description. As a result, object $N$ is described by the following sequence:  $\lbrace <OBJN>F_{N}^{2d}, F_{N}^{v}, F_{Nk_1}^{e}, F_{Nk_2}^{e} \rbrace$. We do not perform the pretraining phase on GT instance segmentation in this experiment. ~\cref{tab:relations-tokens} shows that the object-centric graph representation using triplets improves the performance of the visual grounding task.


We include additional experimental results from ablation studies on scene captioning and visual question answering tasks in the Supplementary Materials.

\section{Conclusion}

In this paper, we propose a new learnable approach to using a 3D semantic scene graph for a large language model to solve 3D vision-language tasks. Detailed experiments demonstrate the effectiveness of this approach, which explicitly takes into account semantic relations between objects represented as 3D point clouds. 
Our method, called 3DGraphLLM, surpasses the baseline approach without semantic relationships on popular ScanRefer, Multi3DRefer, Scan2Cap, ScanQA, and SQA3D datasets. Moreover, 3DGraphLLM achieves state-of-the-art performance in the object grounding task, matching the quality of methods that require five times more inference time.

A limitation of the method is a significant increase in resource consumption with an increase in the edge number for each graph node. At the same time, we showed that taking into account only two edges for each object demonstrates an acceptable trade-off between performance and model quality. 

For further development of the work, it seems appropriate to search for methods to reduce token usage for encoding object relationships in our graph representation. Another important aspect for further work is the creation of methods for generating semantic relations between objects that are robust to imperfections in the instance segmentation of the scene point cloud.

\newpage

\section*{Acknowledgments}
The study was supported by the Ministry of Economic Development of the Russian Federation (agreement with MIPT No. 139-15-2025-013, dated June 20, 2025, IGK 000000C313925P4B0002). 

{
    \small
    \bibliographystyle{ieeenat_fullname}
    \bibliography{main}
}

\newpage

\appendix

\section{Ablation Study. Number of Nearest Neighbors}

We investigate how increasing the number of nearest neighbors affects the quality of scene description and question answering tasks. The number of nearest neighbors varies from 0 to 4.

When no nearest neighbors are used, the model operates as a baseline Chat-Scene approach, representing the scene as a list of objects without semantic relationships. The maximum number of nearest neighbors that can be added for each object is four, constrained by GPU memory limitations during model training. In these experiments, we use GT instance segmentation to eliminate errors in graph construction that could otherwise impact performance on 3D vision-language tasks. The base LLM used in this study is LLAMA3-8B-Instruct.

\begin{figure}
  \begin{center}
  \includegraphics[width=0.8\linewidth]{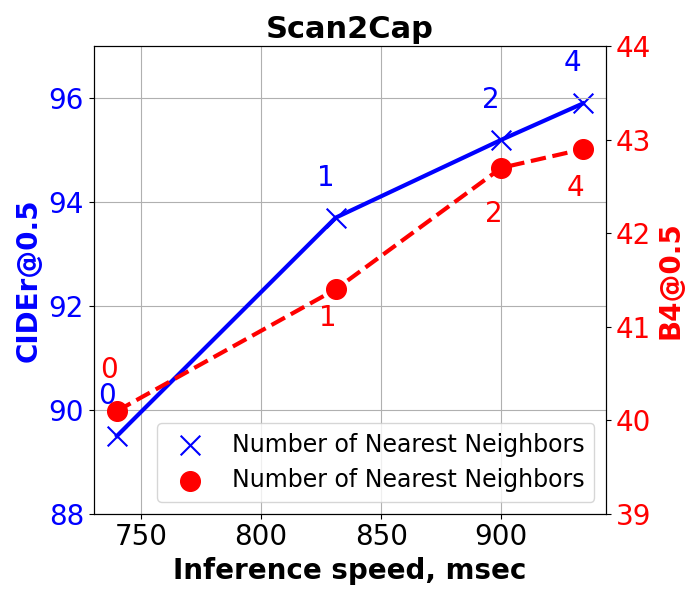}
  \end{center}
  \caption{\label{nn-ablation-scan2cap}
  Dependence of inference speed and dense scene captioning quality on the number of nearest neighbors in the object subgraph. This experiment utilizes the GT instance segmentation.
  }
\end{figure}

As shown in ~\cref{nn-ablation-scan2cap}, increasing the number of nearest neighbors improves the quality of object descriptions in the scene, while causing only a slight increase in generation time. Notably, in dense scene captioning tasks, inference speed depends not only on the number of tokens used to represent the scene but also on the length of the generated description. Therefore, we observe only a small increase in generation time when comparing graphs with two and four nearest neighbors.

\begin{figure}
  \begin{center}
  \includegraphics[width=0.8\linewidth]{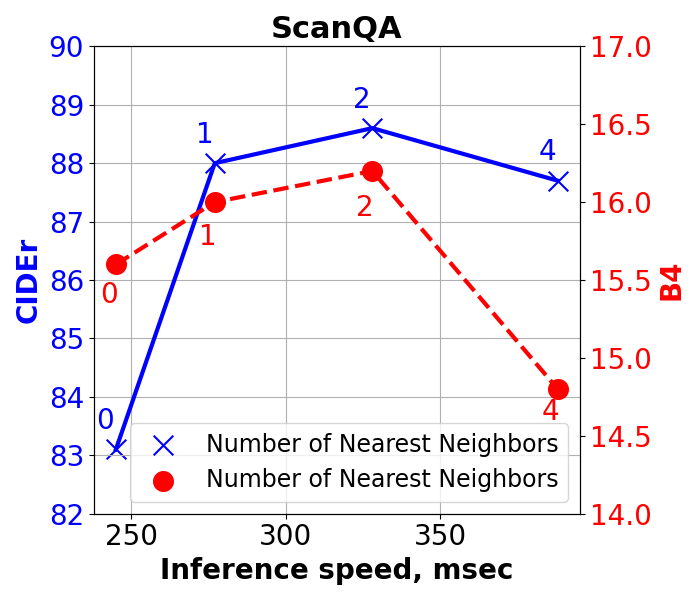}
  \end{center}
  \caption{\label{nn-ablation-scanqa}
  Dependence of inference speed and question answering quality on ScanQA dataset on the number of nearest neighbors in the object subgraph. This experiment utilizes the GT instance segmentation.
  }
\end{figure}

\begin{figure}
  \begin{center}
  \includegraphics[width=0.8\linewidth]{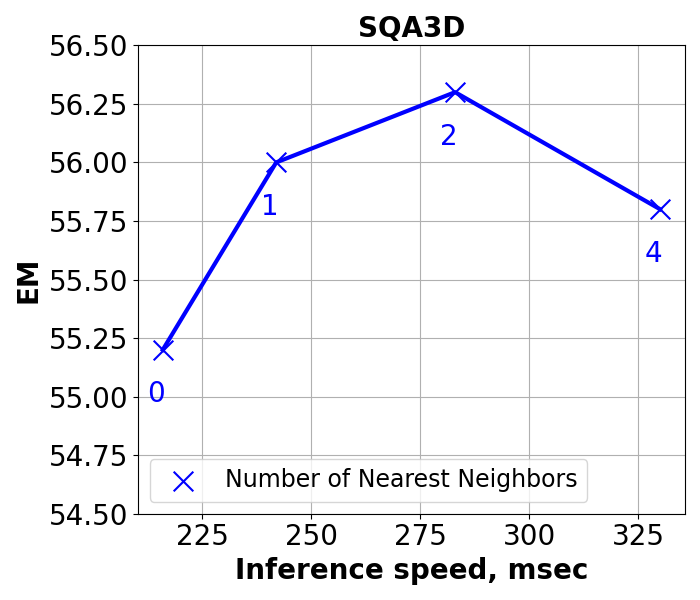}
  \end{center}
  \caption{\label{nn-ablation-sqa3d}
  Dependence of inference speed and question answering quality on SQA3D dataset on the number of nearest neighbors in the object subgraph. This experiment utilizes the GT instance segmentation.
  }
\end{figure}

\cref{nn-ablation-scanqa} and \cref{nn-ablation-sqa3d} show that semantic relationships between objects enhance performance in question answering tasks on the ScanQA and SQA3D datasets. However, for question answering tasks, the optimal number of nearest neighbors is 2, as increasing it to 4 leads to a drop in performance. The impact of semantic edges is harder to assess in question answering tasks than in visual grounding or object description tasks, since some question types do not require knowledge of object spatial relationships.

For further experiments, we select 2 nearest neighbors, as it provides the best trade-off between performance gains and computational complexity across all three tasks.

\begin{table*}[t]
\scriptsize

\centering

\begin{tabular}{llllllllllll}
    \hline
      \multicolumn{1}{c}{} &
      \multicolumn{1}{l}{Instance  } &
      \multicolumn{1}{l}{Number } &
      \multicolumn{1}{l}{Minimal } &
      \multicolumn{1}{c}{ScanRefer} &
      \multicolumn{1}{c}{Multi3DRefer} &
      \multicolumn{2}{c}{Scan2Cap} &
      \multicolumn{2}{c}{ScanQA} &
      \multicolumn{1}{c}{Sqa3D} \\
    Methods & segmentation &  of edges &  distance, cm & Acc@0.5\textuparrow & F1@0.5\textuparrow & C@0.5\textuparrow & B-4@0.5\textuparrow  & C\textuparrow & B-4\textuparrow & EM\textuparrow \\
    \hline
    3DGraphLLM-0 & GT & 0 & - & 61.5 & 64.4 & 89.5 & 40.1 & 83.1 & 15.6 & 55.2 \\
    3DGraphLLM-2 & GT & 2 & 0 & \textbf{66.9} & \textbf{69.9} & \textbf{95.2} & \textbf{42.7} & \textbf{88.6} & \textbf{16.2} & \textbf{56.3} \\
    \hline
    3DGraphLLM-0 & Mask3D & 0 & - & 52.0 & 55.1 & 80.0 & 37.5 & 84.0 & \underline{15.8} & 53.8  \\
    3DGraphLLM-2 & Mask3D & 2 & 0 & 55.6 & 58.2 & 80.8 & 36.4 & \underline{85.7} & 15.1 & \underline{56.0} \\
    3DGraphLLM-2 & Mask3D (+ NMS) & 2 & 0 & \underline{55.7} & \underline{58.6} & \underline{82.3} & \underline{36.8} & \textbf{86.2} & \textbf{16.0} & \textbf{56.2} \\
    3DGraphLLM-2 & Mask3D (+ NMS) & 2 & 1 & \textbf{56.2} & \textbf{58.7} & \textbf{82.9} & \textbf{37.3} & 85.4 & 15.1 & 55.6 \\
    \hline
    3DGraphLLM-0 &  OneFormer3D & 0 & - & 50.0 & 52.8 & \textbf{73.5} & \textbf{34.3} & \textbf{87.3} & \textbf{16.5} & 53.8 \\
    3DGraphLLM-2 &  OneFormer3D & 2 & 0 & \underline{52.8} & \underline{55.8} & 70.2 & 32.7 & \underline{83.3} & \underline{15.0} & \textbf{55.0} \\
    3DGraphLLM-2 & OneFormer3D (+NMS) & 2 & 1  & \textbf{54.6} & \textbf{57.2} & \underline{72.4} & \underline{33.0} & 81.3 & 12.9 & \textbf{55.0} \\
  \hline
\end{tabular}
\caption{Ablation study on semantic edge role depending on quality of instance segmentation.}
\label{tab:app:instance-segm}
\end{table*}

\section{Ablation Study. Quality of Instance Segmentation.}

For the graph, where the vertices contain objects obtained from 3D instance segmentation, we observe a consistent improvement in the performance of all three 3D Vision-Language tasks. When transitioning from GT segmentation to a noisy graph composed of vertices obtained using Mask3D, we also observe improvements in metrics for all three tasks (see ~\cref{tab:app:instance-segm}). However, this improvement is less pronounced compared to GT instance segmentation.

We compare the graphs of nearest neighbors obtained from GT instance segmentation and Mask3D instance segmentation. The analysis shows that Mask3D instance segmentation contains a large number of object duplicates, as the scene is always divided into N=100 segments. The presence of object duplicates among the neighbors leads to a reduction in useful information about the object's environment in its subgraph. To address the duplicates in the vertices of the subgraphs, we use two filters. The NMS filter with an IoU threshold of 0.99 removes duplicate objects from the neighbors. The minimum distance filter between the centers of the object point clouds excludes the object's own duplicates from its neighbors.

~\cref{tab:app:instance-segm} shows that adding these filters consistently improves the performance of visual grounding and object description tasks for the graph obtained through Mask3D instance segmentation. Since we expect an effect from adding semantic edges specifically for these tasks, we keep this filter in further experiments.

We also experiment with different methods for instance segmentation to create scene graph vertices. We use another method for instance segmentation, OneFormer3D, filtering out vertices with confidence $< 0.1$. We observe that for scene graphs with such vertices, semantic edges improve the performance of the visual grounding task. At the same time, the combination of nearest-neighbor filters proves effective for this type of scene segmentation, increasing the performance of object grounding and scene captioning tasks. However, since OneFormer3D showed worse results with these hyperparameters compared to Mask3D, and other baseline methods use Mask3D for object proposals, we chose Mask3D for the final version of the pipeline.



\section{Ablation Study. Subgraph Representation.}


We explore the possibility of further flattening the graph by replacing relationship triplets with a sequence of semantic edges.
As a result, object $N$ is described by the following sequence: $\lbrace <OBJN>F_{N}^{2d}, F_{N}^{v}, F_{Nk_1}^{e}, F_{Nk_2}^{e} \rbrace$.

\begin{table*}[!ht]
\scriptsize

  \centering
\begin{tabular}{llllllllllll}
    \hline
      \multicolumn{1}{c}{} &
      \multicolumn{1}{l}{Instance} &
      \multicolumn{1}{l}{Relations } &
      \multicolumn{1}{c}{Number } &
      \multicolumn{1}{c}{ScanRefer} &
      \multicolumn{1}{c}{Multi3DRef} &
      \multicolumn{2}{c}{Scan2Cap} &
      \multicolumn{2}{c}{ScanQA} &
    \multicolumn{1}{c}{Sqa3D} \\
     Methods & segmentation & as triplets & of edges & Acc@0.5\textuparrow & F1@0.5\textuparrow & C@0.5\textuparrow & B-4@0.5\textuparrow  & C\textuparrow & B-4\textuparrow & EM\textuparrow \\
    \hline
    3DGraphLLM-0 & Mask3D & \xmark & 0 & 52.0 & 55.1 & 80.0 & 37.5 & 84.0 & \underline{15.8} & 53.8 \\
    3DGraphLLM-2 & Mask3D & \xmark & 2 & \underline{54.2} & \underline{56.3} & \textbf{87.2} & \underline{39.3} & \underline{85.6} & 15.1 & \textbf{54.6}  \\
    3DGraphLLM-2 & Mask3D & \cmark & 2 & \textbf{54.3} & \textbf{57.3} & \underline{85.6} & \textbf{39.6} & \textbf{87.4} & 14.9 & \underline{54.5}  \\

  \hline
\end{tabular}
\caption{Ablation study on subgraph representation.}

\label{tab:relations-tokens}
\end{table*}

\begin{figure*}[!ht]
  \centering
  \includegraphics[width=0.7\linewidth]{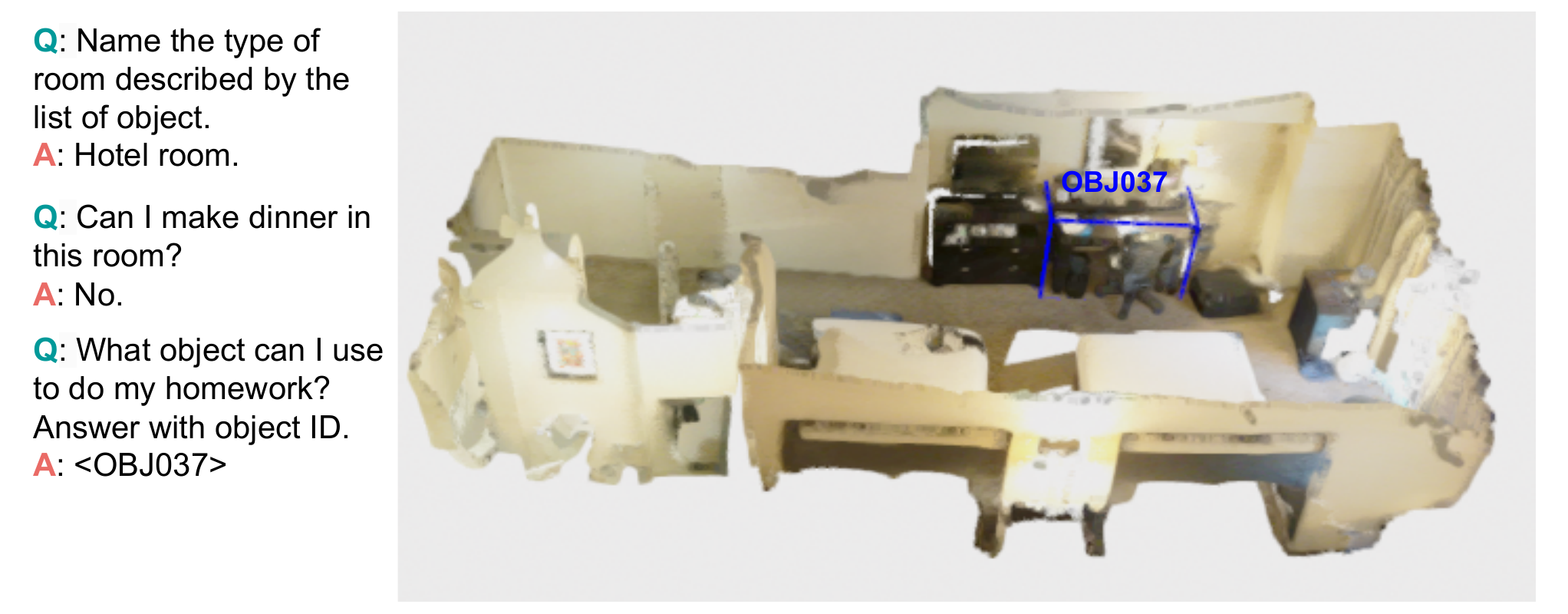}
  \caption{\label{Fig:functional}
Functional queries about the room and objects to the 3DGraphLLM. 3DGraphLLM is capable of answering questions about the functional properties of the room and its room type as well as discerning the functional properties of objects in a room.}
\end{figure*}
\begin{figure*}[!ht]
  \centering
  \includegraphics[width=0.8\linewidth]{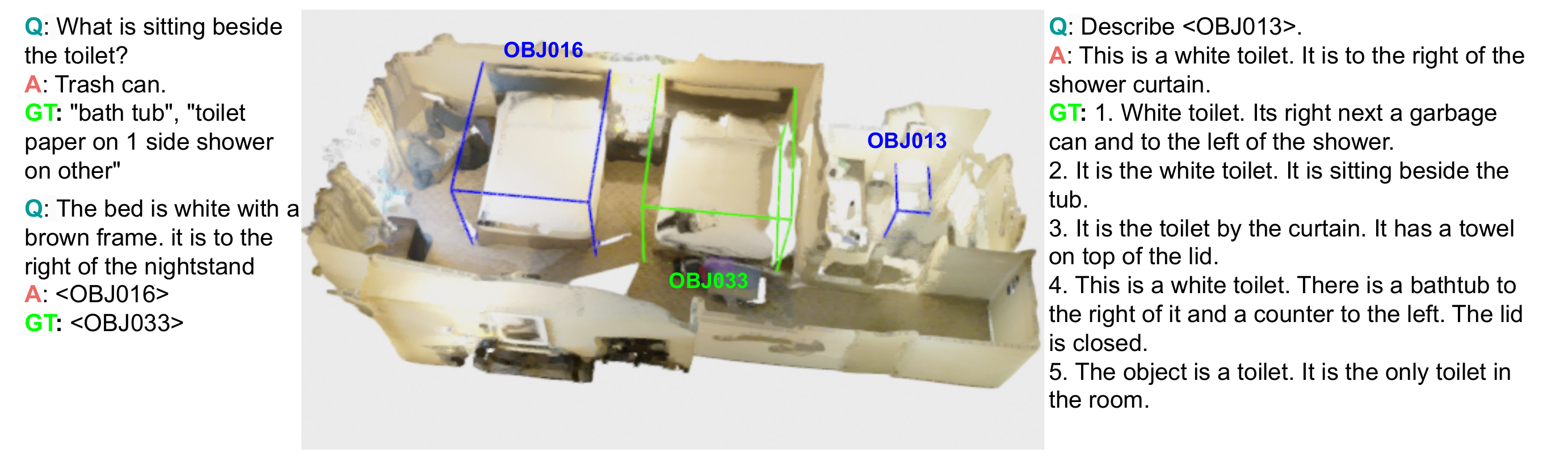}
  \caption{\label{Fig:failure}
Common failure cases of 3DGraphLLM related to spatial relationships. In the question answering task, 3DGraphLLM incorrectly identifies the front/back and left/right directions relative to the observer. In the visual grounding task, 3DGraphLLM confuses left and right. The GT object is highlighted in green, and the 3DGraphLLM prediction is highlighted in red. In the object captioning task, 3DGraphLLM uses a spatial reference not mentioned in the GT descriptions. However, the description is correct qualitatively.}
\end{figure*}

 \cref{tab:relations-tokens} shows that representing relationships as triplets improves the performance of the object grounding task. 
The object-centered representation of relationships improves question answering performance on ScanQA according to the CIDEr metric, while the BLEU-4 metric remains on par with the flat graph representation as a sequence of relationships. For the SQA3D dataset, both approaches yield comparable results. In the dense scene captioning task, we observe a decrease in the CIDEr metric but an improvement in the Scan2Cap metric. However, as shown in ~\cref{sec:failure}, n-gram-based metrics may produce unreliable results when evaluating text generated by LLMs. Considering this, and the fact that the impact of semantic edges is most interpretable in the object grounding task, we represent relationships as triplets in subsequent experiments.

\section{Functional Queries}

We illustrate the ability of 3DGraphLLM to leverage common sense knowledge in its responses to question types not present in the training dataset in ~\cref{Fig:functional}.

\section{Common Failure Cases}
\label{sec:failure}

We illustrate the most common failure cases of 3DGraphLLM related to spatial relationships in ~\cref{Fig:failure}. 


It is important to note that the quality metrics in the Scan2Cap, ScanQA, and SQA3D benchmarks are based on n-gram-based metrics comparing generated answers with reference ones, such as BLEU-1, BLEU-2, BLEU-3, BLEU-4, CIDEr, ROUGE-L, METEOR. The Exact Matching (EM) metric compares the exact match of the answer with the GT answer.  The drawback of these metrics is that if an object description or answer to a question contains spatial relationships not present in the reference descriptions, it leads to a decrease in the score. Additionally, these metrics are unable to adequately evaluate LLM responses, considering the richness of formulations and the freedom to choose visual and spatial properties of an object that may be mentioned by the model. These responses represent a special type of "failure cases," illustrated in ~\cref{Fig:failure} on the right. For this type, the object description or answer to the question is correct from a qualitative point of view but shows zero value according to the metric CIDEr. 

\section{Scalability with Number of Scene Objects}
\label{sec:number-of-objects}
To evaluate scalability, we analyze how memory usage and inference time vary with the number of objects in a scene.   The maximum number of objects considered corresponds to the highest counts observed in the ScanNet and 3RScan datasets. As shown in~\cref{tab:inference_metrics}, both memory usage and inference time increase gradually as the number of objects in the scene grows. This demonstrates that while resource consumption scales with object count, the growth remains manageable and does not compromise the method’s practical applicability within real-world scene configurations.

\begin{table}[h!]
\centering
\scriptsize
\begin{tabular}{ccc}
\hline
\textbf{Number of Objects} & \textbf{Memory Usage (Gb)} & \textbf{Inference Time for 1 token (s)} \\
\hline
10 & 23 & 0.08 \\
50 & 28 & 0.14 \\
100 & 35 & 0.23 \\
\hline
\end{tabular}
\caption{Performance metrics for varying number of objects in a 3D scene.}
\label{tab:inference_metrics}
\end{table}

\end{document}